%% file: LatentRansac.tex
\documentclass[10pt,twocolumn,letterpaper]{article}

\usepackage{cvpr}
\usepackage{times}
\usepackage{epsfig}
\usepackage{graphicx}
\usepackage{amsmath}
\usepackage{amssymb}

\usepackage[pagebackref=true,breaklinks=true,letterpaper=true,colorlinks,bookmarks=false]{hyperref}

\cvprfinalcopy 

\def\cvprPaperID{3535} 

\ifcvprfinal\fi

\input{./0_defs.tex}

\begin{document}

\title{Latent RANSAC}

\author{Simon Korman \\
Weizmann Institute of Science, Israel\\
\and
Roee Litman\\ 
General Motors, Israel\\
}

\maketitle

\begin{abstract}

We present a method that can evaluate a RANSAC hypothesis in constant time, i.e. independent of the size of the data.
A key observation here is that correct hypotheses are tightly clustered together in the latent parameter domain.
In a manner similar to the generalized Hough transform we seek to find this cluster, only that we need as few as \emph{two} votes for a successful detection.
Rapidly locating such pairs of similar hypotheses is made possible by adapting the recent "Random Grids" range-search technique.
We only perform the usual (costly) hypothesis verification stage upon the discovery of a close pair of hypotheses. We show that this event rarely happens for incorrect hypotheses, enabling a significant speedup of the RANSAC pipeline.

The suggested approach is applied and tested on three robust estimation problems: camera localization, 3D rigid alignment and 2D-homography estimation. We perform rigorous testing on both synthetic and real datasets, demonstrating an improvement in efficiency without a compromise in accuracy. Furthermore, we achieve state-of-the-art 3D alignment results on the challenging ``Redwood'' loop-closure challenge.

\end{abstract}

\input{./10_intro.tex}
\input{./20_method.tex}

\input{./30_results.tex}

\input{./40_conclu.tex}

{

\input{./LatentRansac.bbl}
}

\end{document}

%% file: 0_defs.tex
\usepackage{overpic}
\usepackage{algorithm2e}
\usepackage{setspace}

\usepackage{sidecap}
\usepackage{amsmath}
\usepackage{amssymb}
\usepackage{epstopdf}
\usepackage{amsthm}
\usepackage{algorithm2e}
\usepackage{bm}
\usepackage{multirow}

\hyperbaseurl{https://}

\makeatletter
\newcommand*{\algotitle}[2]{%
  \stepcounter{algocf}%
  \hypertarget{algocf.title.\theHalgocf}{}%
  \NR@gettitle{#1}%
  \label{#2}%
  \addtocounter{algocf}{-1}%
}
\makeatother

\graphicspath{{./figures/}}

%% file: 10_intro.tex
\section{Introduction}

Despite the recent success of (deep-) learning based methods in computer vision, numerous applications still use ``old-fashioned'' robust estimation methods for model fitting, such as RANSAC \cite{fischler1981random}.
This is especially true for problems of a strong geometric nature such as image alignment, camera localization and $3$D reconstruction.
Robust estimation methods of these types largely follow the ``hypothesize and test" paradigm which has strong roots in statistics, and are highly attractive due to their ability to fit a model to data that is highly corrupted with outliers. Additionally, they have been successfully applied to many problems in computer vision and robotics achieving real time performance.

As an example, in the field of image (or shape) alignment, novel features and descriptors have been introduced to facilitate matching, including ones that are learned. 
However, once these features are matched, for a parametric model to be fitted, robust estimation methods like RANSAC are used to cope with corrupted sets of putative matches.

Geometric models that are commonly amenable to such a robust estimation process include: $2$D-homography, camera localization, the essential and fundamental matrices that describe epipolar constraints between images, rigid $3$D motion and more.

\begin{figure}[t!]
	\addtolength{\tabcolsep}{0pt}
	\centering
	\begin{tabular}{cc}
		\includegraphics[width=0.38\columnwidth]{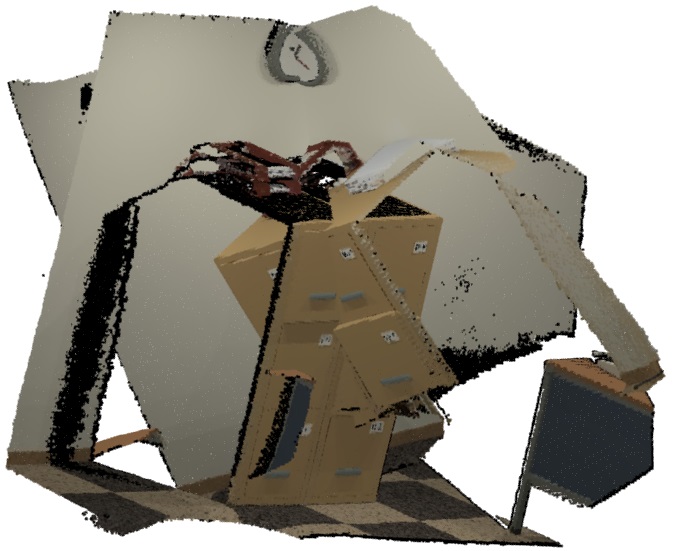} &
		\includegraphics[width=0.535\columnwidth]{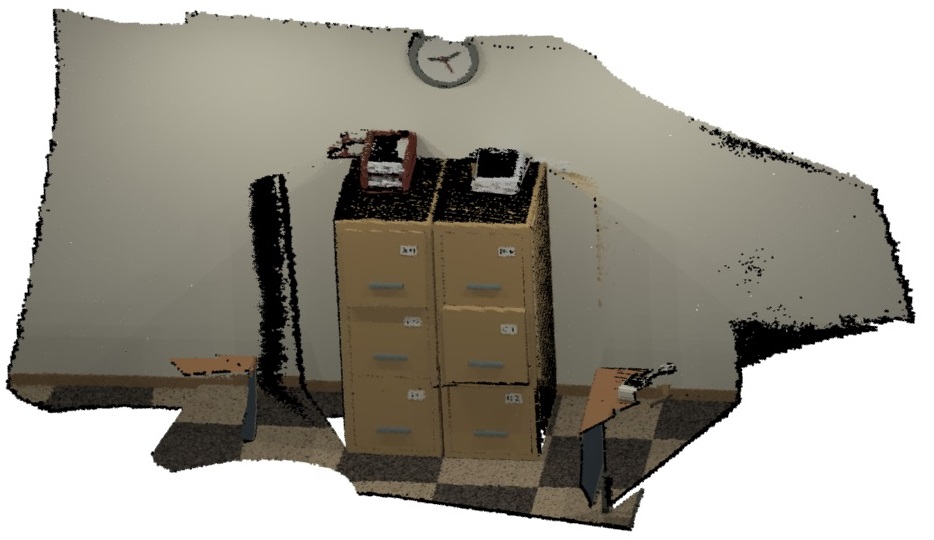}
	\end{tabular}
	\caption{\label{fig:redwood_teaser}
		\textbf{3D alignment result} of two methods on a pair of fragments from the ``Redwood'' dataset \cite{choi2015robust}.
		Our method (right) produces a correct alignment, even though the putative matches contain a mere $2\%$ of inliers, which is very challenging.
		On the left, we see a failure case of the method from \cite{zhou2016fast}, even though it manages to increase inlier rate up to $7\%$.
		Runtimes for this example are $74$ms for our method, and $163$ms for \cite{zhou2016fast}. Table~\ref{table:additional_detailed_examples} shows representative examples for the other problems we handle - PnP and 2D-Homography estimation.
	}\vspace{-9pt}
\end{figure}

\subsection{Background and prior art}

\emph{Consensus maximization} has proven a useful robust estimation approach to solving a wide variety of fitting and alignment problems in computer vision.

Research in this field can be broadly divided into \emph{global} and \emph{local} optimization methods.
Global methods \cite{olsson2008polynomial,zheng2011deterministically,chin2015efficient,Campbell_2017_ICCV} use different strategies to explore the entire solution space enjoy the advantage of having a deterministic nature. Our method, however, belongs to the family of local methods which are typically extremely fast randomized algorithms, potentially equipped with probabilistic success guarantees.

While the proposed method is presented in the context of RANSAC, it is closely related-to and inspired-by other works in the field, such as Hough voting.
We cover these topics briefly.

\begin{figure*}[t!]
	\vspace{-3pt}
	\centering
	\includegraphics[width=0.75\textwidth]{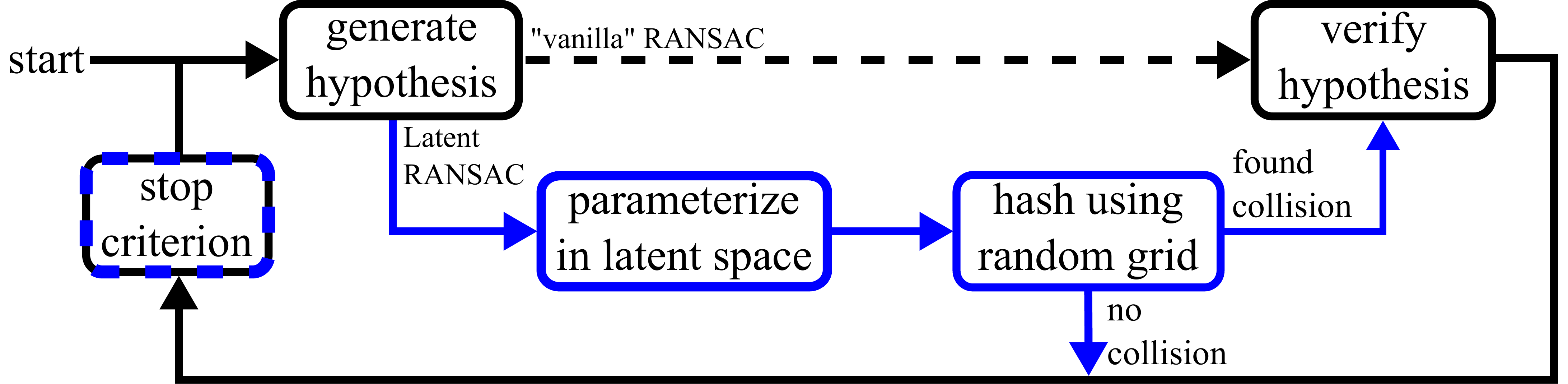}
	\caption{\label{fig:flow_diag}
		\textbf{A flow chart of RANSAC compared to the suggested method.}
		We propose an alternative flow (in blue), in which after a hypothesis is generated it first undergoes a hash procedure, and only verified if a (valid) collision is detected.
		The stop criterion has to be modified as well, to ensure a \emph{second} good hypothesis is drawn with high probability. 
	}\vspace{-8pt}
\end{figure*}

\paragraph*{RANdom SAmple Consensus (RANSAC) \cite{fischler1981random}} is one of the de-facto golden standards for solving robust estimation problems in a practical manner.
Under this paradigm, the space of solutions is explored by repeatedly selecting random minimal subsets of a set of given measurements (e.g. putative matches), for which a model hypothesis is fitted.
These hypotheses are verified by counting the measurements that agree with them up to a predefined tolerance.
This process is repeated until a desired probability to draw \emph{at-least} one pure set of inliers is achieved.

The RANSAC paradigm is a very studied topic, and many methods were suggested to accelerate sampling \cite{tordoff2002guided,chum2005matching,fragoso2013evsac,fragoso2017ansac}, improve stability \cite{chum2003locally, lebeda2012fixing}, or even estimate the tolerance parameter \cite{choi2009starsac,raguram2011recon}.
Some of these extensions are covered in a recent comprehensive survey by Raguram \etal \cite{raguram2013usac}. This survey also suggests USAC -- a framework that combines some RANSAC extensions, yielding excellent results on a variety of problems in terms of accuracy, efficiency, and stability.

While the previous extensions can be seamlessly applied along with the suggested method, the following extension is similar in nature to ours in that it aims to speed up the verification step, but it does so in a very different manner. The Sequential Probability Ratio Test (SPRT) \cite{chum2008optimal} extension was selected in USAC out of several similar methods \cite{capel2005effective,chum2008optimal}
SPRT is based on Wald's sequential test \cite{wald1973sequential}. It attempts to reject a ``bad'' model with high probability, after inspecting only a small fraction of the measurements. While the test is theoretically solid, it relies on two parameters that are assumed to be known a priori, and in practice need to be adaptively estimated at runtime. It is reported to have achieved an improvement of 20\% in evaluation time compared to the simpler bail-out test \cite{capel2005effective}.

\paragraph{Generalized Hough transform (GHT)}
originated from an algorithm for line detection in images \cite{hough1959machine}, which was later generalized to handle arbitrary shapes \cite{duda1972use,ballard1981generalizing}.
The key idea behind this method is that partial observations of the model are casted as votes into a (quantized) solution space, in which the object can be detected as a mode (the location with the most votes).
In practice, GHT has not been shown to scale well to solution spaces of high dimensionality (i.e. higher than $3$), and typically requires \textit{numerous} votes for a mode to be accurately detected.

\paragraph{Between RANSAC and GHT.}
Some works bare resemblance to both of the mentioned approaches.
Our work can be seen as one of these: While it fits naturally into the RANSAC pipeline, it has some similarities to GHT in the sense that it seeks to find the mode in the parameter domain, only that it needs as few as \emph{two} votes to detect it.

A method by Den-Hollander et al. \cite{den2007combined} also lays somewhere between RANSAC and GHT: To increase the probability of obtaining a pure set of inlier matches, a sub-minimal set is drawn. The remaining degrees of freedom (DoF) are resolved using a voting scheme in a low-dimensional setting. As with all Hough-like methods, an adequate parameterization of the remaining DoF is required. The authors of \cite{den2007combined} provide such a parameterization for the problem of fundamental matrix estimation.

Our method bares a strong resemblance to the ``Randomized Hough Transform'' (RHT) \cite{xu1990new} of Xu et al. in that a vote is
casted into a single cell in the solution domain, generated from a randomly selected minimal set.
However, unlike \cite{xu1990new}, we deal with a hypothesis in \textit{constant time and space}, rather than logarithmic, thanks to the Random Grid hashing mechanism that we adapt.
In addition, while RHT deals with robust curve-fitting (of up to 3 dimensions), we successfully apply our method on a variety of problem domains of higher dimensionality (up to 8 dimensions).

\subsection{Contributions}

The main novelty of the presented method is its ability to handle RANSAC hypotheses in \emph{constant time}, regardless of the number of measurements (e.g. matches).
We show that it is beneficial to handle hypotheses in the latent space, due to an efficient parametrization and hashing scheme that we devise, which can quickly filter candidate hypotheses until a pair of correct ones are drawn.
While this approach comes at the expense of a \textit{small} increase in the number of hypotheses to be examined, it allows for a significant speedup of the RANSAC pipeline.

The new proposed modifications to RANSAC are accompanied by a rigorous analysis which results in an updated stopping criterion and a well understood overall probability of success.
Finally, we validate our method using challenging data in the problems of 2D-homography estimation, 2D-3D based camera localization and rigid-3D alignment, showing state-of-the-art results.

%% file: 20_method.tex
\section{Method}

The `vanilla' RANSAC pipeline can be divided into three main components: hypothesis generation, hypothesis verification and the adaptive stopping mechanism. 
The proposed Latent-RANSAC hypothesis handling fits naturally into the aforementioned pipeline, as can be seen in Figure~\ref{fig:flow_diag}, highlighted in blue.
The additional modules we propose act as a `filter' that avoids the need to verify the vast majority of generated hypotheses:
Instead of verifying each hypothesis by applying it on all of the matches 
(a costly process that takes time linear in the number of matches), we check in \emph{constant time} if a previously generated hypothesis `collides' with the current one, i.e. whether they are close enough (in a sense that will be clarified below).
Only the very few hypotheses that pass this filtering stage progress to the verification stage for further processing.
As a result of the proposed change, the RANSAC stopping criterion needs to modified to guarantee a probability of encountering a second good hypothesis rather than just one.

\paragraph*{Outline}
We begin by covering the $3$ key components of our method: parametrization of the solution space (Section~\ref{sec.parametrization}), Random Grids hashing (Section~\ref{sec.random_grids}) and the modified stopping criterion (Section~\ref{sec.stopping.crit}).
We conclude this part of the paper in Section~\ref{sec.analysis}, with an analysis of our Random Grids hashing process.

\paragraph*{Preliminary definitions}
\label{sec:preliminary}

In our setup, the goal is to robustly fit a geometric model (transform) to a set of matches (correspondences), w.l.o.g. in Euclidean space, where a \emph{match} $\textbf{m}=(\textbf{p},\textbf{q})$ is an ordered pair of points $\textbf{p}\in \mathbb{R}^d$ and $\textbf{q}\in \mathbb{R}^{d'}$.
For a geometric transform $f:\mathbb{R}^d\rightarrow \mathbb{R}^{d'}$ and match $\textbf{m}=(\textbf{p},\textbf{q})$ the \emph{residual error} of the match $\textbf{m}$ with respect to $f$ is the Euclidean distance in $\mathbb{R}^{d'}$ given by:
\begin{equation}
\mathrm{err}(f,\textbf{m}) = \|\textbf{q} - f(\textbf{p})\|.
\end{equation}
Given a set of matches $M=\{\textbf{m}_i\}$ and a tolerance $t$, the \emph{inlier rate} achieved by a transform $f$ is defined as the fraction of matches $\textbf{m}_i\in M$ for which  $\mathrm{err}(f,\textbf{m}_i)\le t$.
We denote the maximal inlier rate for a match-set $M$ by $\omega$. 

\subsection{Parametrization of the solution domain} \label{sec.parametrization}

In the RANSAC pipeline, matches are used \textit{both} for the generation of hypothesis candidates, as well as for their screening.
Since our approach performs the majority of the screening according to some 'similarity' in the space of transformations, we seek a \emph{parametrization} of the transformation space in which distances between transformations can be defined explicitly.
More formally, we define such a parametrization by an \emph{embedding} hypotheses into some $\lambda$-dimensional space $\mathbb{R}^\lambda$, which we call the \emph{latent} space\footnote{the latent space dimension $\lambda$ typically being the number of degrees of freedom of the transformation space.}. We consider the distance between transformations to be given by the $\ell_\infty$ metric between the embedded (or latent) vectors ($\lambda$-tuples). 
Our goal is to use an embedding in which the \emph{distance} between any pair of hypotheses $f_1$ and $f_2$ is tightly related to the \emph{difference} in the way these hypotheses act on matches in the source domain, i.e. to the difference in magnitudes of their residual errors on the matches.
Ideally, for \emph{any} set of matches $M$, 
\begin{equation}\label{eq:laternt_dist}
	\|f_1 - f_2\|_\infty \propto \max_{\textbf{m} \in M} |\mathrm{err}(f_1,\textbf{m}) - \mathrm{err}(f_2,\textbf{m})|.
\end{equation}

\paragraph{2D homography.}

\begin{figure}[h]
	\centering
	\begin{overpic}[width=\columnwidth]{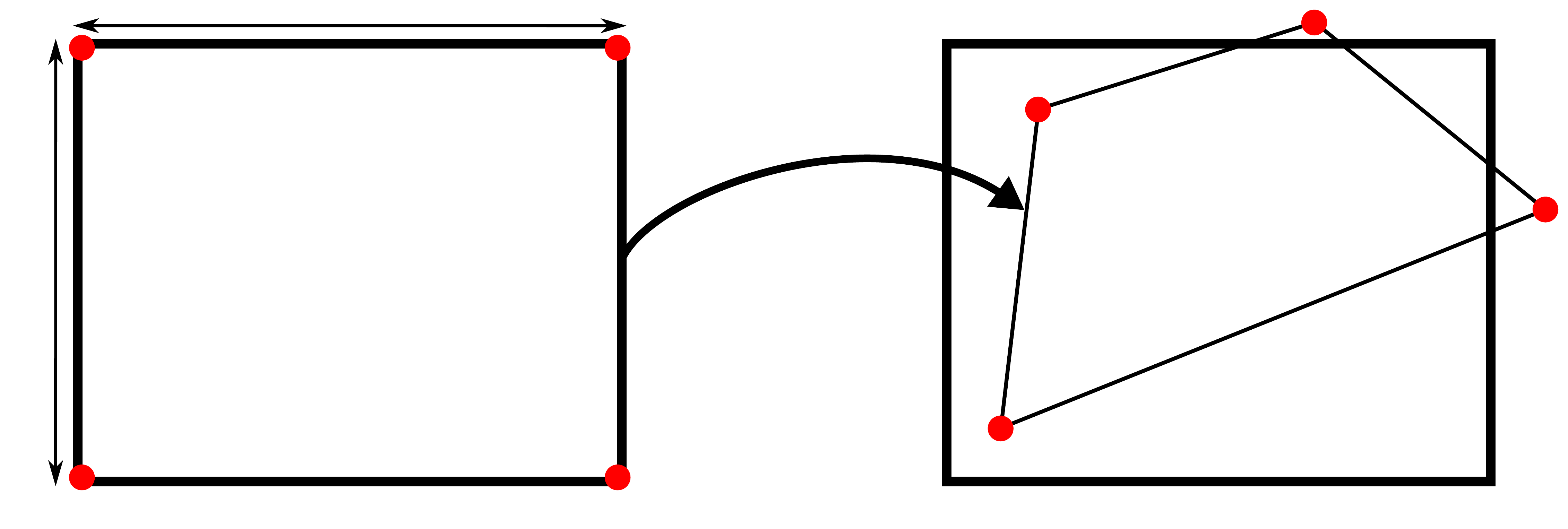}
		\put(09,-2){source image}
		\put(67,-2){target image}
		\put(00,16){$h$}
		\put(20,34){$w$}
		\put(50,17){$H$}
		\put(66,06){$(x_1,y_1)$}
		\put(53,27){$(x_2,y_2)$}		
		\put(83,34){$(x_3,y_3)$}		
		\put(88,15){$(x_4,y_4)$}
		\put(06,05){$(0,0)$}
		\put(06,26){$(h,0)$}		
		\put(28,26){$(h,w)$}		
		\put(28,05){$(0,w)$}
	\end{overpic}
	\vspace{-1ex}	
	\caption{ \label{fig:homog_hash}
		\textbf{Illustration of the 4pt homography parametrization \cite{baker2006parameterizing}.} A homography $H$ is represented in the latent space by mapping the location of the four corners of the source image onto the target image, resulting in the 8-tuple $\textbf{v}_H=(x_1,y_1,\ldots,x_4,y_4)$.
	} \vspace{-5pt}
\end{figure}
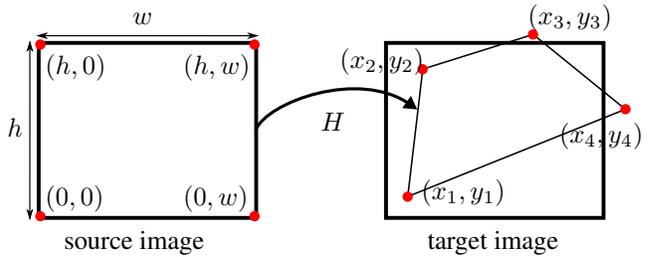

We describe here the parameterization we use for the space of $2$D homographies, which are given by projective matrices in $\mathbb{R}^{3 \times 3}$.
Following previous works (e.g. \cite{litman2015inverting,detone2016deep}), we use the 4pt parameterization~\cite{baker2006parameterizing} that represents a $2$D-homography $H\in\mathbb{R}^{3 \times 3}$ by an 8-tuple $\textbf{v}_H$, defined by the coordinates in the target image that are the result of applying $H$ on the four corners of the source image, as illustrated in Figure~\ref{fig:homog_hash}.

As was noted in \cite{litman2015inverting}, this parametrization has the key property that the difference between match errors of two well-behaved homographies is bounded by the $\ell_\infty$ distance between their $4$pt representations.

\paragraph{The special Euclidean group $SE(3)$,} used to describe rigid motion in $\mathbb{R}^3$, will be used here to solve the problems of Perspective-n-Point (PnP) estimation and Rigid $3$D alignment.
We follow a parametrization that was suggested and used in a line of works of Li et al. \cite{yang2013goicp,Campbell_2017_ICCV}. 
The group $SE(3)$ can be described as the product between two sub-groups $SE(3)=SO(3)\times\mathbb{R}^3$, namely $3$D translations and the special orthogonal group ($3$D rotations).
Each of these 3-dimensional sub-groups is parameterized as a 3-tuple, resulting in a 6-tuple representation $(r,t)$ defined as follows:
The axis-angle vector (3-tuple) $r$ represents the $3$D rotation matrix given by $R_r=\exp([r]_x)$, where $\exp(\cdot)$ is the matrix exponential and $[\cdot]_x$ denotes the skew-symmetric matrix representation. Such vectors $r$ reside in the radius-$\pi$ ball that is contained in the $3$D cube $[-\pi,\pi]^3$. The translation 3-tuple $t$ is a vector in the cube $[-\xi,\xi]^3$ that contains the relevant bounded range of translations for a large enough $\xi$.

Similar to the case of the $2$D-homography parametrization, it is proved in \cite{yang2013goicp} that the difference between match errors of two rigid motions is bounded by the $\ell_2$ distance between their parametrization.

\subsection{Random Grids hashing} \label{sec.random_grids}
Given such a embedding of a generated hypotheses, the heart of our method boils down to a nearest neighbor query search of the current vector through all vectors representing previously generated hypotheses.
More precisely, the task needed to be performed is a range search query for vectors that are at a distance of up to a certain tolerance $t$.

A recent work of Aiger et al. \cite{aiger2013random} turns out to be extremely suitable for this task. They propose Random Grids - a randomized hashing method based on a very simple idea of imposing randomly shifted `grids' over the vector space, checking for vectors that `collide' in a common cell.
The Random Grids algorithm is very fast, and simple to implement - even in comparison with the closely related LSH-based algorithms \cite{andoni2006near}, since the grid is axis aligned and it is uniform (consists of cells in $\mathbb{R}^d$ with equal side length).
Most important, and essential for the speed of our method, is that the range search is done in constant time (i.e. it does not depend on the number of vectors searched against), as opposed to the RANSAC hypothesis validation that requires applying the model and measuring errors on (typically hundreds of) point matches or even the logarithmic-time solution proposed in RHT \cite{xu1990new}.

\paragraph{Hashing scheme.} We are given a representation of the transform $f$ as a vector $\textbf{v}\in \mathbb{R}^\lambda$ (for a $\lambda$-dimensional parameterization).
In the Random Grids \cite{aiger2013random} setting, we hash $\textbf{v}$ into $L$ hash tables $\{T_i\}_{i=1}^L$, each associated with an independent random grid, which is defined by a uniform random shift $O_i\sim U([0,c]^\lambda)$, where $c$ is the cell side length and $\lambda$ is the dimension of the latent vector $\textbf{v}$.
The cell index for $\textbf{v}$ in the table $T_i$ is obtained by concatenating the integer vector $\textbf{z}_i=\left\lfloor\frac{\textbf{v}+O_i}{c}\right\rfloor$ into a single scalar (where $\lfloor\cdot\rfloor$ means ``floor'' operation).
The entire hashing process - initialization, insertion and collision checking, is given in detail in Algorithm~\ref{alg:filtering}.

\begin{algorithm}[t]

\SetAlgoLined
\SetKwInput{Input}{input}\SetKwInput{Output}{output}

\hrulefill

\Input{ (incremental) A candidate transform (matrix) $\!f$}

\textbf{parameters:} number of tables $L$;$\;$ tolerance $t$;$\;$ cell dim. $c$;$\;$ parametrization dim. $\lambda$;
\vspace{-5pt}

\hrulefill

\hspace{-0pt}\parbox{.93\columnwidth}{

\vspace{-4pt}
\textbf{\underline{initialization}:}\vspace{4pt}\\
\ForEach{$i=1,\ldots,L$}
{
\vspace{4pt}
1. Initialize an empty hash table $T_i$.\\ 
2. Randomize offset $O_i\sim U([0,c]^\lambda)$ \vspace{-0pt}
}

\vspace{5pt}
\textbf{\underline{insertion and collision check}} for hypothesis $f$:\vspace{4pt}\\
\ForEach{$i=1,\ldots,L$}
{
\vspace{4pt}
1. Let $\textbf{v}$ be the embedding of $f$ \vspace{4pt}\\
2. The hash index for $\textbf{v}$ is: $\tau_{\textbf{v}}=hash\left(\left\lfloor\frac{\textbf{v}+O_i}{c}\right\rfloor\right)\;$\vspace{4pt} \\
3. If the cell $T_i[\tau_{\textbf{v}}]$ is occupied by a vector $\textbf{u}$, report a collision of $f$ if 
$\|\textbf{v}-\textbf{u}\|_\infty<t$
\vspace{4pt} \\%
4. Store $\textbf{v}$ in $T_i[\tau_{\textbf{v}}]$ \vspace{2pt}
}
}

\hrulefill

\medskip
\caption{Latent-RANSAC hypothesis handling.  \label{alg:filtering}  }
\end{algorithm}


\subsection{Latent-RANSAC stopping criterion} \label{sec.stopping.crit}

The classical analysis of RANSAC provides a simple formula for the number of iterations $n$ required to reach a certain success probability $p_0$ (e.g. 0.99). It is based on the assumption that it is sufficient to have a single `good' iteration in which a pure set of inliers is drawn. Note that this assumption is made for the simplicity of the analysis and is only theoretical, since it ignores e.g. the presence of inlier noise and several possible degeneracies in the data.

Formally, let $G_n$ be the random variable that counts the number of such good iterations out of $n$ attempts. For a minimal set of size $\gamma$ and data with inlier rate of $\omega$, it holds that
\vspace{-4pt}
\begin{equation} \label{eq.P.S.ransac}
p_0=1-P[G_n=0]=1-(1-p)^n
\end{equation}
where $p=\omega^\gamma$. The number of iterations $n$ required to guarantee a desired success probability $p_0$  is therefore:
\begin{equation} \label{eq.needed.iterations.ransac}
n=\frac{\log(1-p_0)}{\log(1-p)}
\end{equation}

\begin{figure}
	\centering
	\includegraphics[width=.9\columnwidth]{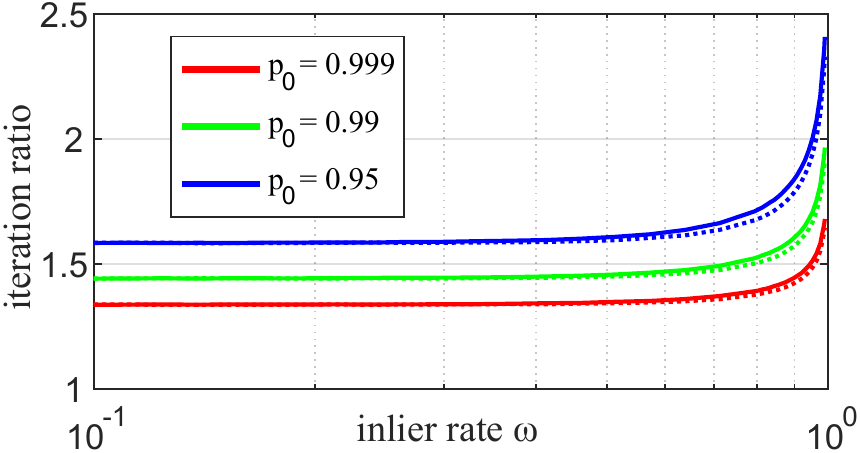}		
	\caption{\label{fig:stopping_ratio}
		\textbf{The ratio between stopping criterions} of Latent-RANSAC \eqref{eq.G.N.geq2} and of RANSAC \eqref{eq.needed.iterations.ransac}. Ratios are shown as a function of inlier rate $\omega$ for several success probabilities (color coded) and for $\gamma$ (minimal sample size) values of 3 (dashed) and 4 (solid). See text for details.}
	\vspace{-10pt}
\end{figure}

A similar simplified analysis can be applied to the Latent-RANSAC scheme. Ignoring the presence of inlier noise, 
the existence of (at least) two `good' iterations is needed for a collision to be detected and the algorithm to succeed.
Therefore, by the binomial distribution we have that
\begin{equation} \label{eq.G.N.geq2}
p_0 = P[G_n\ge 2] = 1-(1-p)^n-n\!\cdot\! p\!\cdot\!(1-p)^{n-1}.
\end{equation}
Based on equations \eqref{eq.needed.iterations.ransac} and \eqref{eq.G.N.geq2}, we plot in Figure~\ref{fig:stopping_ratio} the \emph{ratio} between the number of required iterations $n$ in the case of Latent-RANSAC versus the case of RANSAC.
The ratio is given as a function of the inlier rate $\omega$, at 3 different success rates $p_0$ (color coded), for the two different cases $\gamma=3$ (e.g. in Rigid $3$D motion estimation) and $\gamma=4$ (e.g. in homography estimation).
Interestingly, the ratio attains a small value (less than 2) for inlier rates below $\omega=0.95$, and converges to small constant values as the inlier rate decreases. The very high inlier rates for which the ratio is large are of no concern, since the absolute number $n$ is extremely low in this range.

In the next section, as part of an analysis of the Random Grid hashing, we derive a more realistic stopping criterion that depends also on the success probability of the Random Grid based collision detection, which clearly depends on the inlier noise level.

\subsection{Random Grid analysis} \label{sec.analysis}

We cover two aspects of Random Grids. First, we extend the stopping criterion from Section~\ref{sec.stopping.crit} to consider the probability that a colliding pair of good hypotheses will be detected. Next, we discuss causes of false collision detection, which can have an affect on the algorithm runtime.

\paragraph{stopping criterion.}
Let $R(i)$ be the event that the random grid component succeeds (detects a collision), given $i$ good iterations out of a total of $n$. We can now update Equation~\eqref{eq.G.N.geq2}, taking this success probability into account:
\begin{equation}\label{eq.Latent.p_0}
p_0=\sum_{i\ge 2}P[G_n=i]\!\cdot\!P[R(i)] \;\ge\; P[G_n\ge 2]\!\cdot\!P[R(2)]
\end{equation}
where the inequality holds due to the fact that $P[R(i)]$ monotonically increases with $i$.

A final lower bound on $p_0$ (from which the stopping criterion is determined) can be obtained by substituting the expression for $P[G_n\ge 2]$ from \eqref{eq.G.N.geq2} into \eqref{eq.Latent.p_0} together with a lower bound on $P[R(2)]$ which we provide next.

Recall that $R(2)$ is the event that the random grid hashing succeeds given that two successful hypotheses were generated. We will, more explicitly, denote this event by $R_L(2)$, for a random grid that uses $L$ hash tables.

The analysis in \cite{aiger2013random} is rather involved since it deals with the Euclidean $\ell_2$ distance.
Using $\ell_\infty$ distances we are able to derive the success probability of finding a true collision in our setup, as a function of the random grid parameters, in a simpler manner.
Assuming a tolerance $t$ in the latent space, determined by (inlier) noise level of the data, using a random grid with cell dimension $c$ and a \emph{single} table results in
\begin{equation} \label{eq.P.RG.simple}
P[R_1(2)] \ge \left( 1-\frac{t}{c} \right) ^ \gamma
\end{equation}
since a pair of pure-inlier transformations (which differ by at most $t$) must share the same independently offsetted bin indices in each of the $\gamma$ dimensions.

Finally, using $L$ hash tables, randomly and independently generated, we obtain:
\begin{equation}\label{eq.P.RG.final}
 P[R(2)]= P[R_L(2)] \ge 1-\left(1-P[R_1(2)]\right)^L
\end{equation}


\paragraph{False collisions.}
We now discuss the expected number of false collisions that are found by the hashing scheme. It is important to understand why false collision might happen, as they have an effect on the overall runtime of our pipeline.

Recall that $n$ is the overall number of iterations of the pipeline, and hence it is also the total number of samples inserted into each hash table.
There are two kinds of false collisions to consider.
The first kind happens due to the fact that the random grid cell size $c$ might be larger than the tolerance $t$. 
Following the recommendation in \cite{aiger2014reporting} we set the cell size $c$ to be not much larger than the tolerance $t$, resulting in a small number of such false collisions.
In any case, this kind of collision has a small impact on the runtime, since it will be filtered by the tolerance test (step 3 in Algorithm~\ref{alg:filtering}) at constant time cost.

The second kind of false collision is one that passes the tolerance test (step 3). 
Since it is not the true model, it is associated with some inlier rate $\zeta$. 
If $\zeta\ll\omega$, the probability of this collision appearing before we have reached the stopping criterion is negligible. 
Empirically, we observe very few (typically less than $15$) collisions that pass the tolerance test up to the stopping of the algorithm. 
These are the only kind of collisions that incur a non-negligible penalty (in runtime only) since they invoke the verification process that every ``vanilla'' RANSAC hypothesis goes through.

%% file: 30_results.tex
\section{Results}

In order to evaluate our method, we performed extensive tests on both real and synthetic data. The Latent-Ransac algorithm is applied to the problems of $2$D-homography estimation (Section~\ref{sec.results.homog}), Perspective-n-Point (PnP) estimation (Section ~\ref{sec.results.pnp}) and Rigid $3$D alignment (Section~\ref{sec.results.rigid3d}). It is compared with USAC, with or without the well known SPRT \cite{chum2008optimal} extension, which is a very different technique for accelerating RANSAC's model verification phase.

\paragraph{Implementation details.}

Our method naturally extends the standard RANSAC pipeline, according to the changes highlighted in Figure~\ref{fig:flow_diag}. Our implementation (for which we use the shorthand LR) extends the excellent C++  implementation USAC~\cite{raguram2013usac}, with the noted changes in the specific modules.
This enables an easy way to compare with a state-of-the-art RANSAC implementation, and allows our method to enjoy the same extensions used by USAC (such as its local optimization component LO-RANSAC~\cite{chum2003locally}). In addition, their implementation includes the SPRT \cite{chum2008optimal} extension, the most commonly used acceleration of RANSAC's model verification phase. We use the shorthand SPRT to refer to USAC using this extension, and RANSAC when not using it.

In addition, we use the OpenGV library \cite{kneip2014opengv} for PnP model fitting (Kneip's P3P algorithm \cite{kneip2011novel}) and for Rigid $3$D model fitting (Arun's algorithm \cite{arun1987least}).
We make our modifications to the code publicly available\footnote{\url{github.com/rlit/LatentRANSAC}}.

The parameters of the proposed method were selected empirically based on synthetic data (disjoint from other experiments), and kept fixed throughout (for details, see \cite{latent_supp}).
Parameters common to all settings: probability of success $p_0=0.99$, number of hash tables $L=4$; Random Grid cell size $c=1.8t$, where the LR tolerance $t$ (and the RANSAC threshold used) are specified separately for each experiment;
Maximal number of iterations $n=5\times10^6$;
Hash table size of $n/10$, resulting in addressing indices of $19$ bits and less than $4$ms initialization time  (see \cite{latent_supp} for hash table implementation details).

\begin{figure}[t]
	\centering
	\begin{overpic}
		[width=1\columnwidth]{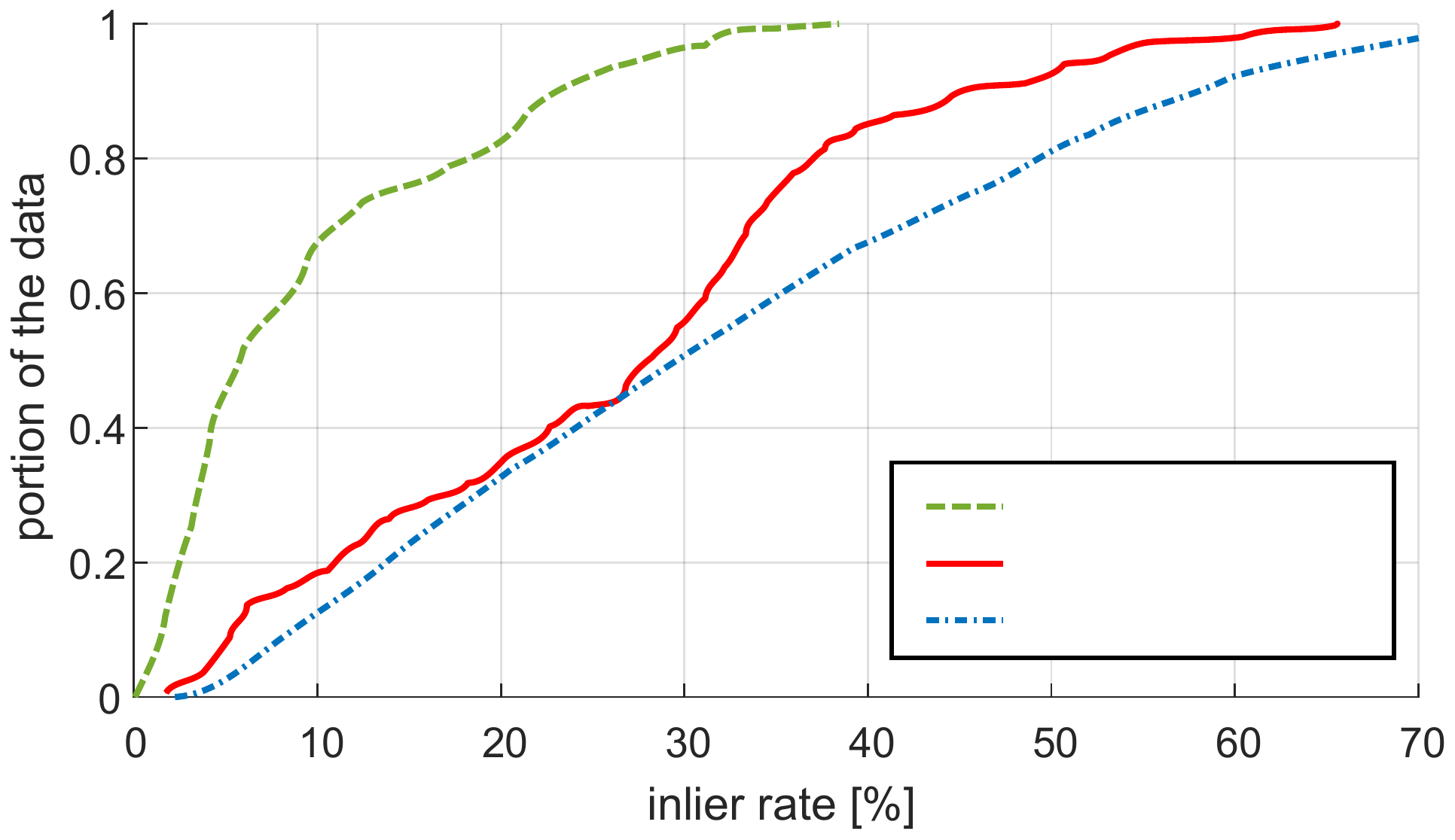}
		\put(71,22){\footnotesize \cite{choi2015robust} Redwood}  %
		\put(71,18){\footnotesize \cite{shao2003zubud} ZuBuD} %
		\put(71,14){\footnotesize \cite{kendall2015posenet} Old Hospital} %
	\end{overpic}
	\caption{\label{fig:ir_cdfs}
		\textbf{Inlier rate cumulative distribution} (CDF) of the three real data sets we use.
		The dashed curve was taken only over Redwood pairs with provided ground truth.
	}
\end{figure}

\subsection{2D-homography estimation} \label{sec.results.homog}

\begin{figure*}[t]
	\centering
	\includegraphics[width=\textwidth]{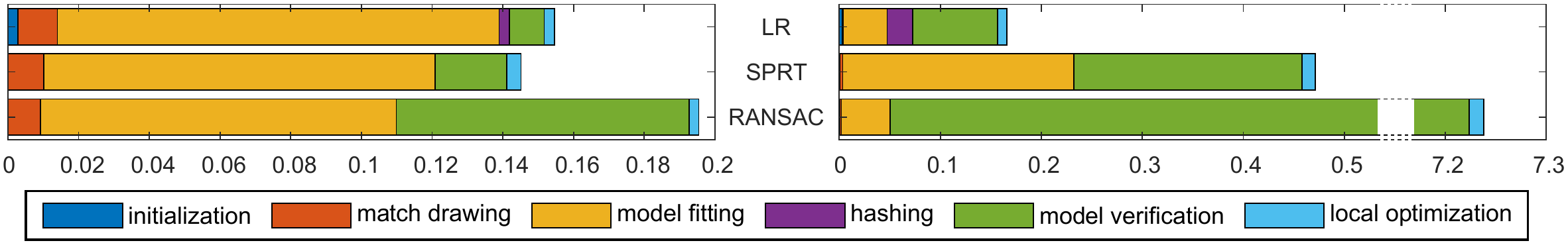}
	\caption{\label{fig:runtime_breakdowns}
		\textbf{Runtime breakdown per pipeline module comparing LR to SPRT and RANSAC.} These are average per-instance runtimes (in seconds) for $2$D-homography estimation (\textbf{left}) and PnP estimation (\textbf{right}) taken over each of the entire data-sets used for the evaluation.} \vspace{-2pt}
\end{figure*}

We create a large body of $2$D-homography estimation instances using the Zurich buildings data-set \cite{shao2003zubud}. The data-set consists of sequences of 5 snapshots taken from different street-level viewpoints for 201 buildings. The images are typically dominated by planar facade structures and hence each pair of images in a sequence is related by a $2$D homography (or perhaps more than one in the case of several planes).

We computed SIFT~\cite{lowe2004distinctive} features for each image and created sets of corresponding features for each of the 10 (ordered) pairs of images in a sequence using the VLFeat library \cite{vedaldi08vlfeat}. Following \cite{raguram2013usac}, we generated ground-truth by running $10^7$ iterations of both RANSAC and LR on each pair, and saved the highest inlier rate detected (along with the resulting homography) as the 'optimal' inlier rate for the image pair. A small set of image pairs (132 out of 2010) with very low inlier-rate was manually removed from the evaluation, since the inlier feature locations did not reside on an actual single plane in the scene, and were \emph{very} noisy.

The 1878 resulting matching instances are challenging: many pairs have low inlier rates, that result from (i) the planar area of interest typically covering only part of each image; (ii) large viewpoint changes; (iii) large presence of repetitive patterns (e.g. windows or pillars). See  Figure \ref{fig:ir_cdfs} for the distribution of inlier rates for this data-set.

We ran RANSAC and SPRT with a threshold of $8$ pixels to capture the hard cases, and following \cite{latent_supp} LR tolerance $t$ was set to $70$ pixels in the latent domain. We ran 100 independent trials of each method and summarize the results in Table \ref{tbl:homography_zurich}. We arrange the image pairs into four groups according to their 'optimal' inlier rate (defined above), and the size of each group is shown at the bottom of the table. For each group we report the average, and $95$-percentile of runtimes for each method. We also report each method's success rate (averaged over all pairs in the group), which is the ratio between the detected inlier-rate and the 'optimal' inlier rate. 

\input{tables/homog_zurich_data_formatted}

As can be seen, SPRT and LR (modestly) accelerate RANSAC at the harder inlier rate ranges, where the overall runtime is longer. LR achieves this with no loss in accuracy, while SPRT fails on some cases in the 0-10 range.

The detailed runtime breakdowns shown in Figure~\ref{fig:runtime_breakdowns} (left) reveals two important points that should be made here. First, in homography estimation, methods that accelerate the RANSAC evaluation stage (i.e. LR and SPRT) have a relatively small potential improvement gap since the runtime of RANSAC-based homography estimation is dominated by the model fitting stage (this is not the case for the other problems we deal with, as will be seen later). Second, the improved acceleration in the lower ranges is significant when considering the overall time taken to fit the entire data-set, since the majority of time is spent on these difficult cases which are surprisingly not rare (12\% and 20\% of all pairs are in the 0-10 and 10-20 ranges respectively).

\subsection{Perspective-n-Point (PnP) estimation} \label{sec.results.pnp}

We chose to use the P3P algorithm \cite{kneip2011novel} for minimal sample model fitting (for all methods) due to its good accuracy-efficiency trade off compared to other alternatives. 

Putative 2D-3D matches were generated following image-to-SfM localization pipeline from \cite{irschara2009structure}. We used images from \cite{kendall2015posenet}, where vocabulary trees and queries were generated based on the train-test split therein. Even though we tested on all scenes from \cite{kendall2015posenet}, result are presented only for the ``Old Hospital'' scene as it contains the best balance of moderate and challenging cases in terms of inlier rates (see Figure~\ref{fig:ir_cdfs}). More result are presented in \cite{latent_supp}

We ran RANSAC and SPRT with a threshold of $2.9^\circ$, and following \cite{latent_supp} LR with a tolerance $t$ of $5^\circ$ (in the latent domain), and the translation-to-angle ratio of the embedding was $2.1\tfrac{cm}{rad}$.

The results are summarized in table \ref{tbl:pnp_oldhospital}. We grouped the 182 query images (PnP instances) by increasing level of difficulty, using the ground truth inlier rates. For each of the 182 query images we ran 10 independent trials and report average and 95th percentile of the detected inlier rates. It can be seen that LR achieves more than an order of magnitude acceleration compared to RANSAC, at a comparable accuracy (as before, the success rate is the ratio between the detected and optimal inlier rates). SPRT achieves similar acceleration factors and accuracy at the higher inlier rates (above $10\%$). It is not as efficient or as accurate the lower range (below $10\%$).

In the PnP problem, the existence of fast fitting algorithms (e.g. \cite{kneip2011novel}), make the verification stage the main time consumer in RANSAC. As can be seen in the time breakdown in Figure~\ref{fig:runtime_breakdowns} (right), the costly verification time (over 95\% of RANSAC time) is practically eliminated by LR.

\input{tables/pnp_old_hospital_formatted}

\subsection{Rigid 3D alignment} \label{sec.results.rigid3d}

To evaluate the Rigid $3$D alignment application of Latent-RANSAC, we use the registration challenge of the recent ``Redwood'' benchmark proposed by Choi \etal \cite{choi2015robust}.
This dataset was generated from four synthetic $3$D scenes, each divided into $52$ point-cloud fragments on average. 
While from synthetic origin, these fragments contain high-frequency noise and low-frequency distortion that simulate scans created by consumer depth cameras.

The challenge is to perform global $3$D registration between every pair of fragments of a given scene, in order to provide candidate pairs for trajectory loop closure.
A correctly `detected' pair is one for which the point clouds overlap by at least $30\%$ and the reported transformation is sufficiently accurate (see \cite{choi2015robust} for details).
The main goal in this benchmark, as stated by \cite{choi2015robust}, is to achieve \emph{high recall} while relying on a post-process to later remove false-matches.

Aside from the benchmark, Choi \etal \cite{choi2015robust} present a simple extension (CZK) to the Point-Cloud-Library (PCL) \cite{holz2015registration} implementation of \cite{rusu2009fast}. The method of CZK showed state-of-the-art performance, while comparing to previous methods like
OpenCV \cite{drost2010model},
4PCS \cite{aiger20084pcs} and its extension
super4PCS \cite{mellado2014super}.
Fast Global Registration (FGR) \cite{zhou2016fast} is a recent novel optimization process presented by Zhou \etal, which achieves an order of magnitude runtime acceleration on this dataset, at a competitive recall-precision performance.
They perform the costly nearest-neighbor (NN) search only once (unlike previous methods which use them in their inner loop), while introducing several fast and simple methods to filter false matches.

We chose to follow \cite{choi2015robust, holz2015registration, zhou2016fast} and feed our framework with putative matches based on FPFH features \cite{rusu2009fast}. Following \cite{latent_supp}, we  use LE tolerance $t$ of $24$cm in the latent space, and angle-to-translation ratio of $3.6\tfrac{rad}{cm}$ for the embedding.
Inspired by FGR we perform the NN search only once. We then only apply one single filter (also used in \cite{choi2015robust, zhou2016fast}), an approximate congruency validation on each minimal sample drawn.

\begin{figure}
	\centering
	\begin{overpic}
		[width=0.45\textwidth]{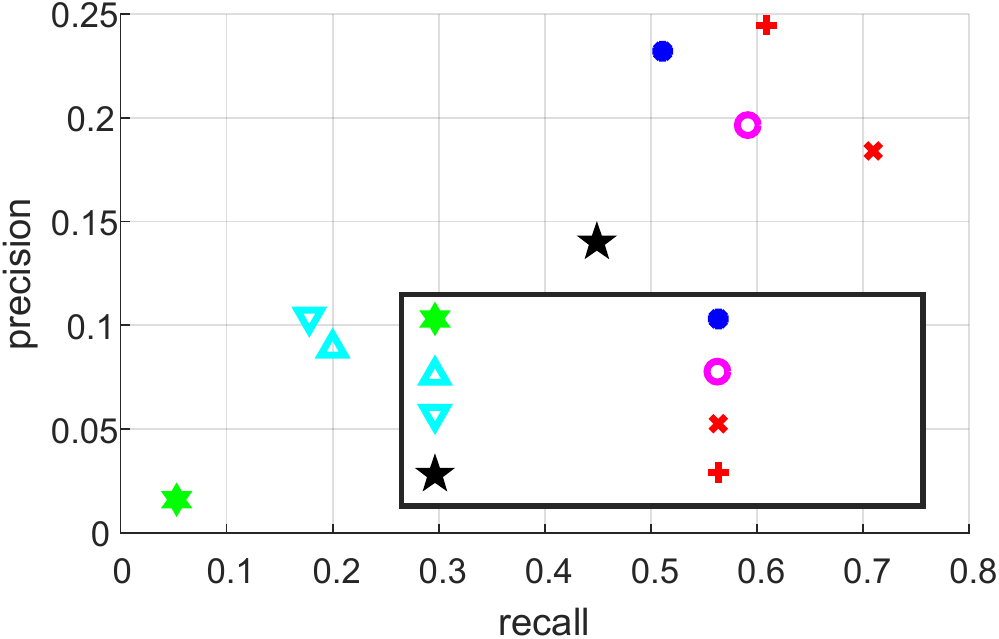}
		\put(46.5,31){\footnotesize \cite{drost2010model} OpenCV}  %
		\put(46.5,25.8){\footnotesize \cite{aiger20084pcs} 4PCS} %
		\put(46.5,20.3){\footnotesize \cite{mellado2014super} super4PCS} %
		\put(46.5,15){\footnotesize \cite{holz2015registration} PCL} %
		\put(75,31){\footnotesize \cite{zhou2016fast} FGR} %
		\put(75,25.8){\footnotesize \cite{choi2015robust} CZK} %
		\put(75,20.3){\footnotesize LR}
		\put(75,15){\footnotesize LR (strict)}
	\end{overpic}
	\caption{\label{fig:redwood_prre}
		\textbf{Performance on the ``redwood'' benchmark~\cite{choi2015robust}.}
		Our method achieves state-of-the-art \emph{recall} in the standard setting (marked by a red `{\tt x}'), while using a stricter threshold (marked by a red `{\tt +}') dominates all previous result in both precision and recall.
		See description in the text for further details.
	}
\end{figure}
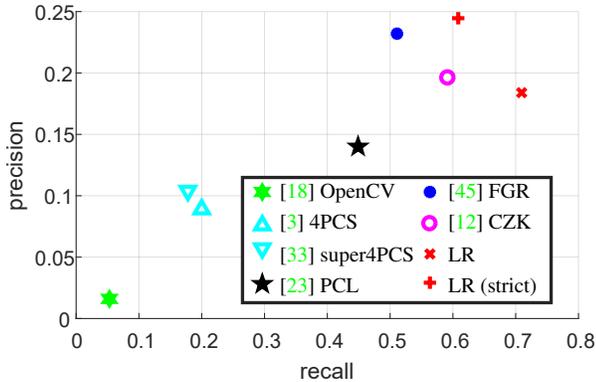

Figure~\ref{fig:redwood_prre} shows a comparison of Latent RANSAC to other results reported in \cite{zhou2016fast}.
Our method clearly achieves the highest recall value (the main goal), at a precision slightly below that of CZK.
Furthermore, we are able to dominate all previous results in precision and recall simultaneously by using a slightly stricter setting, when reporting only pairs with overlap of over $40\%$ rather than $30\%$.

We attribute our high performance mainly to the fact that we perform almost no filtering to the putative matches, such as the bidirectional search and tuple filtering done in \cite{zhou2016fast}, normal-agreement in \cite{choi2015robust,holz2015registration} or the drawing of a non-minimal set of 4 matches in \cite{choi2015robust}.
Using this ``naive'' nearest neighbor FPFH feature matching avoids filtering of true correspondences (enabling higher recall), but this comes at the cost of some very low inlier-rates, as can be seen in Figure~\ref{fig:ir_cdfs}. Our algorithm is able to deal with such inlier rates successfully (and efficiently), as was shown in the other experiments of this section and in Figure~\ref{fig:redwood_teaser}.

Another attractive property of our method in this benchmark is its runtime, presented in Table~\ref{tbl:rigid_times}. Our runtime is close to that of FGR, which we outperform significantly in terms of recall. Note, however, that our method is actually faster than FGR whenever the inlier rate is above $5\%$, as the number of iterations given by \eqref{eq.G.N.geq2} is very low.
Verification consumes a considerable part of the runtime of methods like \cite{choi2015robust,holz2015registration}, while we perform the costly overlap verification only upon the detection of a collision ($13.3$ times per run on average). Additionally, we perform overlap calculation only once as done in FGR.

\begin{table}[h!]
	\hspace{-5pt}
\addtolength{\tabcolsep}{-1pt}
	\begin{tabular}{ccccc}
		\hline
		\textbf{method} & PCL~\cite{holz2015registration} & CZK~\cite{choi2015robust} & FGR~\cite{zhou2016fast} & LR \\
		\hline
		\textbf{avg. time} (sec) \hspace{-5pt} & $3.8$ & $7.5$ & $\textbf{0.21}$ & $0.40$ \\
		\hline
	\end{tabular}
    \vspace{4pt}
	\caption{\label{tbl:rigid_times}
		\textbf{Average runtimes on the ``redwood dataset''}, excluding normals and FPFH \cite{rusu2009fast} calculation time which are $24$ms and $300$ms on average, respectively.
		A breakdown of our method's timing includes $84$ms for feature matching, $305$ms for the latent RANSAC pipeline, and $13$ms for overlap calculation.
	}	\vspace{-6pt}
\end{table}

%% file: tables/homog_zurich_data_formatted.tex
\begin{table}[h] \vspace{-7pt}
\centering
\addtolength{\tabcolsep}{-2.8pt}{\small
\begin{tabular}{|c|c|c|c|c|c|}
\cline{3-6}
\multicolumn{1}{r}{}&&\multicolumn{4}{|c|}{\small{\textbf{inlier rate range (in \%)}}}\\
\hline
\small{\textbf{\:measure\:\:}}&\small{\textbf{method}}&\small{ 0-10 }&\small{ 10-20 } &\small{ 20-40 } &\small{ 40-100 } \\
\hline
\hline
\multirow{3}{*}{\parbox[c][][c]{1.3cm}{\centering \textbf{runtime} \\ \footnotesize{avg. (95\%) } \\ (millisecs) }}& {\footnotesize{RANSAC}}
& \footnotesize{1,490 (6,594)} & \footnotesize{35 (97)} & \footnotesize{5 (10)} & \footnotesize{\textbf{6} (\textbf{9})}   \vspace{-1pt}\\
 \cline{2-2}
& {\footnotesize{SPRT}}
& \footnotesize{\textbf{1,129} (4,659)} & \footnotesize{\textbf{23} (\textbf{62})} & \footnotesize{\textbf{4} (\textbf{7})} & \footnotesize{\textbf{6} (\textbf{9})}   \vspace{-1pt}\\
 \cline{2-2}
& {\footnotesize{LR}}
& \footnotesize{1,209 (\textbf{4,626})} & \footnotesize{32 (85)} & \footnotesize{7 (11)} & \footnotesize{9 (12)}   \vspace{-1pt}\\
 \cline{2-2}
\hline
\hline
\multirow{3}{*}{\textbf{success}}& {\footnotesize{RANSAC}}
& \footnotesize{\textbf{93.39}\% } & \footnotesize{95.53\% } & \footnotesize{96.28\% } & \footnotesize{97.33\% }   \vspace{-1pt}\\
 \cline{2-2}
& {\footnotesize{SPRT}}
& \footnotesize{88.57\% } & \footnotesize{95.71\% } & \footnotesize{96.27\% } & \footnotesize{97.48\% }   \vspace{-1pt}\\
 \cline{2-2}
& {\footnotesize{LR}}
& \footnotesize{93.07\% } & \footnotesize{\textbf{95.88}\% } & \footnotesize{\textbf{96.55}\% } & \footnotesize{\textbf{97.59}\% }   \vspace{-1pt}\\
 \cline{2-2}
\hline
\hline
\multicolumn{2}{|c|}{\textbf{\# of instances}}& \footnotesize{234} & \footnotesize{378} & \footnotesize{655} & \footnotesize{611}   \vspace{-1pt}\\
 \cline{2-2}
\hline
\end{tabular}}\vspace{4pt}
  \caption{\textbf{$2$D Homography fitting on Zurich Buildings \cite{shao2003zubud}.} Best results are shown in bold. See text for further details.} \vspace{-1pt}
  \label{tbl:homography_zurich}
\end{table}

%% file: tables/pnp_old_hospital_formatted.tex
\begin{table}[h] \vspace{-7pt}
\centering
\addtolength{\tabcolsep}{-4.7pt}{\small
\begin{tabular}{|c|c|c|c|c|c|}
\cline{3-6}
\multicolumn{1}{r}{}&&\multicolumn{4}{|c|}{\small{\textbf{inlier rate range (in \%)}}}\\
\hline
\small{\textbf{\:measure\:\:}}&\small{\textbf{method}}&\small{ 0-10 }&\small{ 10-20 } &\small{ 20-40 } &\small{ 40-100 } \\
\hline
\hline
\multirow{3}{*}{\parbox[c][][c]{1.3cm}{\centering \textbf{runtime} \\ \footnotesize{avg. (95\%) } \\ (millisecs) }}& {\footnotesize{RANSAC}}
& \footnotesize{4.2e4 (1.7e5)} & \footnotesize{2,336 (5,347)} & \footnotesize{190 (471)} & \footnotesize{41 (71)}   \vspace{-1pt}\\
 \cline{2-2}
& {\footnotesize{SPRT}}
& \footnotesize{2,760 (1.7e4)} & \footnotesize{40 (\textbf{71})} & \footnotesize{15 (20)} & \footnotesize{\textbf{12} (16)}   \vspace{-1pt}\\
 \cline{2-2}
& {\footnotesize{LR}}
& \footnotesize{\textbf{913} (\textbf{4,403})} & \footnotesize{\textbf{39} (72)} & \footnotesize{\textbf{14} (\textbf{18})} & \footnotesize{\textbf{12} (\textbf{15})}   \vspace{-1pt}\\
 \cline{2-2}
\hline
\hline
\multirow{3}{*}{\textbf{success}}& {\footnotesize{RANSAC}}
& \footnotesize{\textbf{95.54}\% } & \footnotesize{98.13\% } & \footnotesize{99.38\% } & \footnotesize{99.12\% }   \vspace{-1pt}\\
 \cline{2-2}
& {\footnotesize{SPRT}}
& \footnotesize{91.94\% } & \footnotesize{\textbf{98.23}\% } & \footnotesize{\textbf{99.40}\% } & \footnotesize{\textbf{99.17}\% }   \vspace{-1pt}\\
 \cline{2-2}
& {\footnotesize{LR}}
& \footnotesize{94.73\% } & \footnotesize{98.14\% } & \footnotesize{99.39\% } & \footnotesize{99.11\% }   \vspace{-1pt}\\
 \cline{2-2}
\hline
\hline
\multicolumn{2}{|c|}{\textbf{\# of instances}}& \footnotesize{29} & \footnotesize{27} & \footnotesize{91} & \footnotesize{35}   \vspace{-1pt}\\
 \cline{2-2}
\hline
\end{tabular}}\vspace{4pt}
  \caption{\textbf{PnP fitting on the OldHospital scene from PoseNet \cite{kendall2015posenet}.} Best results are shown in bold. See text for further details.} \vspace{-1pt}
  \label{tbl:pnp_oldhospital}
\end{table}

%% file: 40_conclu.tex
\section{Future work}
In this work we presented Latent-RANSAC: a novel speed-up of the hypothesis handling stage of the RANSAC pipeline. We have shown its advantages on challenging matching problems, that include very low inlier rates, in the domains of homography estimation, camera localization and rigid 3D motion estimation.

Latent-RANSAC has the potential to be extended to additional domains. Of particular interest is finding an appropriate parametrization of the more challenging fundamental matrix domain, which is classically tackled using RANSAC.

The good results that Latent-RANSAC achieves on the "Redwood" benchmark come to show the advantage of being able to handle highly corrupted "raw"s data (over 60\% of the fragment pairs have under 10\% inlier rate). This is since the alternative of filtering the data to reduce the rate of outliers comes at the risk of loss of informative data. The challenge, however, remains to do so efficiently, especially for search spaces of high dimensionality.

\begin{table}[h!]
	\centering
	\addtolength{\tabcolsep}{-4.98pt}
	\begin{tabular}{|c|c|c|c|c|}
		\hline
		\emph{instance} & \emph{measure} & \small{USAC} & \small{SPRT} & \small{LR} \\
		\hline
		%
		\hline
		\small{\textbf{`\emph{Old Hospital}'}} & \small{inlier rate (\%)} & \footnotesize{2.9$\pm$0.1} & \footnotesize{0.0$\pm$0.0} & \footnotesize{2.9$\pm$0.1} \\
		\small{\textbf{seq 8 frame 12}} & \small{Sampson err.} & \footnotesize{0.27$\pm$0.16} & \footnotesize{failed} & \footnotesize{0.025$\pm$0.12} \\
		\small{\#matches:} \footnotesize{9,917} & \small{\#samples} & \footnotesize{.19$\pm$.012} & \footnotesize{5.0$\pm$0} & \footnotesize{.25$\pm$.034} \\
		\multirow{3}{*}{\includegraphics[width = 0.28\linewidth]{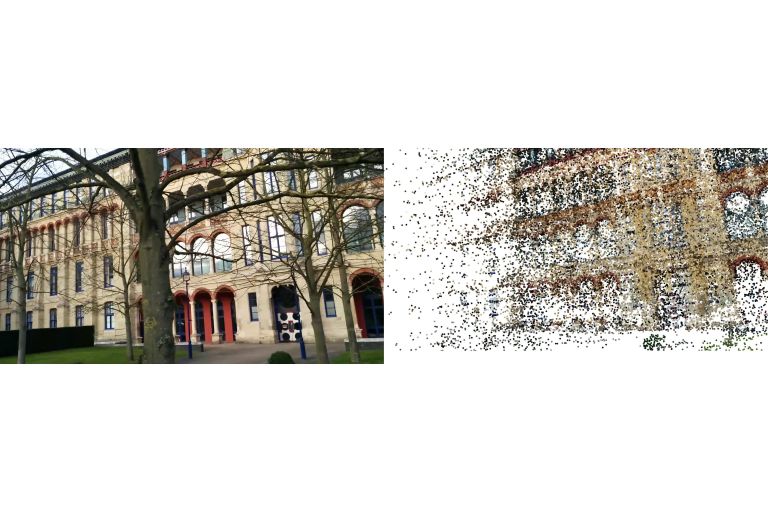}} & \small{\#fitting} & \footnotesize{.19$\pm$.011} & \footnotesize{5.0$\pm$0} & \footnotesize{.26$\pm$.034} \\
		& \small{\#verification} & \footnotesize{.62$\pm$.037} & \footnotesize{16.4$\pm$.002} & \footnotesize{.011$\pm$.002} \\
		& \small{runtime [sec]} & \footnotesize{271.0$\pm$20.8} & \footnotesize{49.8$\pm$2.3} & \footnotesize{6.9$\pm$1.3} \\
		\hline
		%
		%
		\hline
		\small{\textbf{`\emph{building 187}'}} & \small{inlier rate (\%)} & \footnotesize{4.2$\pm$0.1} & \footnotesize{4.2$\pm$1.3} & \footnotesize{4.2$\pm$0.1} \\
		\small{\textbf{views 3,5}} & \small{Sampson err.} & \footnotesize{0.4$\pm$0.2} & \footnotesize{1.8$\pm$1.9} & \footnotesize{0.6$\pm$0.0} \\
		\small{\#matches:} \footnotesize{622} & \small{\#samples} & \footnotesize{1.9$\pm$.31} & \footnotesize{2.7$\pm$.93} & \footnotesize{2.7$\pm$.21} \\
		\multirow{3}{*}{\includegraphics[width = 0.28\linewidth]{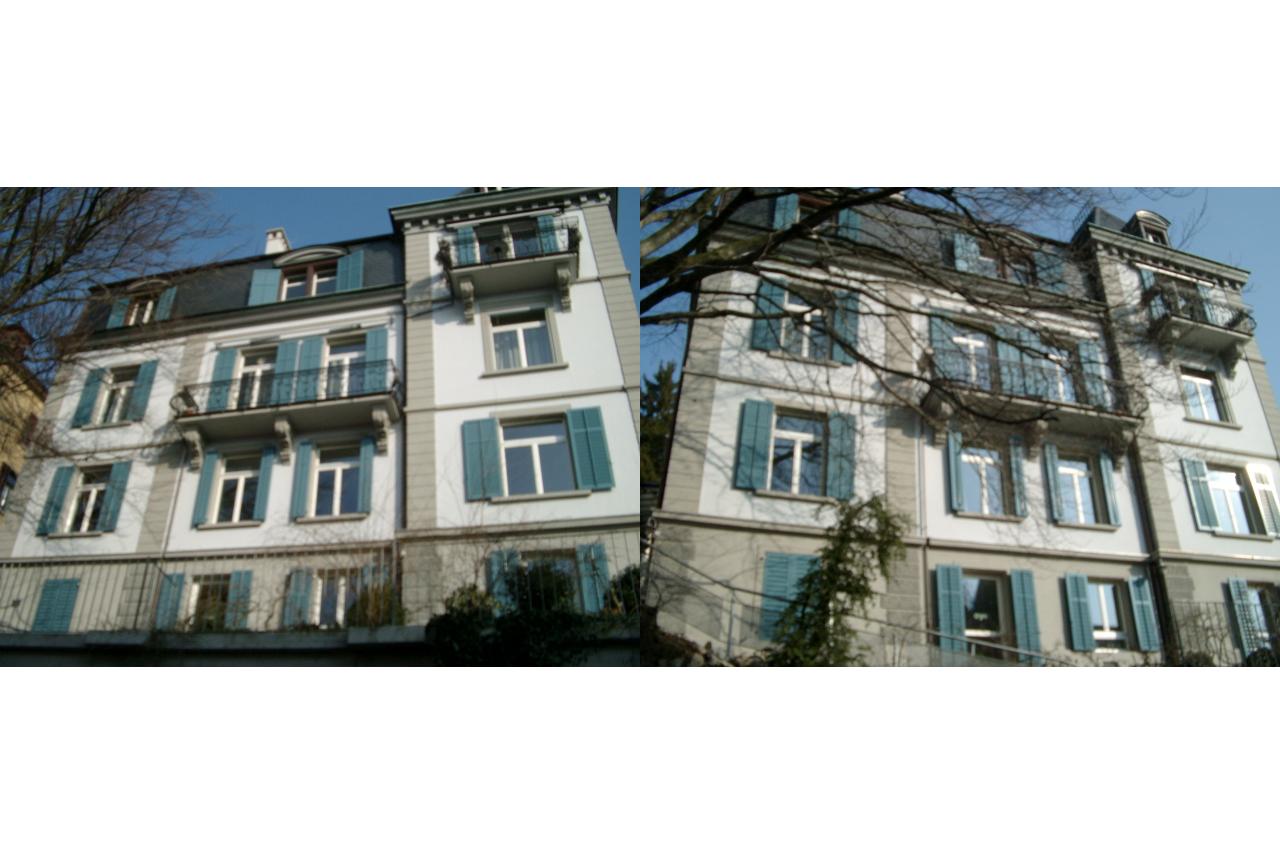}} & \small{\#fitting} & \footnotesize{.19$\pm$.030} & \footnotesize{.27$\pm$.092} & \footnotesize{.26$\pm$.021} \\
		& \small{\#verification} & \footnotesize{.19$\pm$.30} & \footnotesize{.27$\pm$.092} & \footnotesize{.025$\pm$.003} \\
		& \small{runtime [sec]} & \footnotesize{2.83$\pm$0.65} & \footnotesize{2.29$\pm$0.88} & \footnotesize{2.24$\pm$0.23} \\
		\hline
	\end{tabular}\vspace{2pt}
	\caption{\label{table:additional_detailed_examples}
		\textbf{Detailed example results.}
		All measures are reported in terms of median $\pm$ std. Numbers of samples as well as fitting and verification invocations are in millions ($10^6$). 
		\textbf{Top}: 10 iterations of PnP estimation in the PoseNet data \cite{kendall2015posenet}, error in radians. 
		\textbf{Bottom}: 100 iterations of Homography estimation in the Zurich Buildings (ZuBuD) \cite{shao2003zubud} data. 
		See Figure \ref{fig:redwood_teaser} for a detailed rigid-3d estimation example.
	}\vspace{-15pt}
\end{table}

%% file: LatentRansac.bbl
\begin{thebibliography}{10}\itemsep=-1pt

\bibitem{aiger2014reporting}
D.~Aiger, H.~Kaplan, and M.~Sharir.
\newblock Reporting neighbors in high-dimensional euclidean space.
\newblock {\em SIAM Journal on Computing}, 43(4):1363--1395, 2014.

\bibitem{aiger2013random}
D.~Aiger, E.~Kokiopoulou, and E.~Rivlin.
\newblock Random grids: Fast approximate nearest neighbors and range searching
  for image search.
\newblock In {\em Proceedings of the IEEE International Conference on Computer
  Vision}, pages 3471--3478, 2013.

\bibitem{aiger20084pcs}
D.~Aiger, N.~J. Mitra, and D.~Cohen-Or.
\newblock 4-points congruent sets for robust pairwise surface registration.
\newblock {\em ACM Transactions on Graphics (TOG)}, 27(3):85, 2008.

\bibitem{andoni2006near}
A.~Andoni and P.~Indyk.
\newblock Near-optimal hashing algorithms for approximate nearest neighbor in
  high dimensions.
\newblock In {\em Foundations of Computer Science, 2006. FOCS'06. 47th Annual
  IEEE Symposium on}, pages 459--468. IEEE, 2006.

\bibitem{arun1987least}
K.~S. Arun, T.~S. Huang, and S.~D. Blostein.
\newblock Least-squares fitting of two 3-d point sets.
\newblock {\em IEEE Transactions on pattern analysis and machine intelligence},
  (5):698--700, 1987.

\bibitem{baker2006parameterizing}
S.~Baker, A.~Datta, and T.~Kanade.
\newblock Parameterizing homographies.
\newblock {\em Technical Report CMU-RI-TR-06-11}, 2006.

\bibitem{ballard1981generalizing}
D.~H. Ballard.
\newblock Generalizing the hough transform to detect arbitrary shapes.
\newblock {\em Pattern recognition}, 13(2):111--122, 1981.

\bibitem{Campbell_2017_ICCV}
D.~Campbell, L.~Petersson, L.~Kneip, and H.~Li.
\newblock Globally-optimal inlier set maximisation for simultaneous camera pose
  and feature correspondence.
\newblock In {\em The IEEE International Conference on Computer Vision (ICCV)},
  Oct 2017.

\bibitem{capel2005effective}
D.~P. Capel.
\newblock An effective bail-out test for ransac consensus scoring.
\newblock In {\em BMVC}, 2005.

\bibitem{chin2015efficient}
T.-J. Chin, P.~Purkait, A.~Eriksson, and D.~Suter.
\newblock Efficient globally optimal consensus maximisation with tree search.
\newblock In {\em Proceedings of the IEEE Conference on Computer Vision and
  Pattern Recognition}, pages 2413--2421, 2015.

\bibitem{choi2009starsac}
J.~Choi and G.~Medioni.
\newblock Starsac: Stable random sample consensus for parameter estimation.
\newblock In {\em Computer Vision and Pattern Recognition, 2009. CVPR 2009.
  IEEE Conference on}, pages 675--682. IEEE, 2009.

\bibitem{choi2015robust}
S.~Choi, Q.-Y. Zhou, and V.~Koltun.
\newblock Robust reconstruction of indoor scenes.
\newblock In {\em IEEE Conference on Computer Vision and Pattern Recognition
  (CVPR)}, 2015.

\bibitem{chum2005matching}
O.~Chum and J.~Matas.
\newblock Matching with prosac-progressive sample consensus.
\newblock In {\em Computer Vision and Pattern Recognition, 2005. CVPR 2005.
  IEEE Computer Society Conference on}, volume~1, pages 220--226. IEEE, 2005.

\bibitem{chum2008optimal}
O.~Chum and J.~Matas.
\newblock Optimal randomized ransac.
\newblock {\em IEEE Transactions on Pattern Analysis and Machine Intelligence},
  30(8):1472--1482, 2008.

\bibitem{chum2003locally}
O.~Chum, J.~Matas, and J.~Kittler.
\newblock Locally optimized ransac.
\newblock In {\em Pattern Recognition}, pages 236--243. Springer, 2003.

\bibitem{den2007combined}
R.~J. Den~Holl and E.~A. Hanjalic.
\newblock A combined ransac-hough transform algorithm for fundamental matrix
  estimation.
\newblock In {\em in 18th British Machine Vision Conference. University of}.
  Citeseer, 2007.

\bibitem{detone2016deep}
D.~DeTone, T.~Malisiewicz, and A.~Rabinovich.
\newblock Deep image homography estimation.
\newblock {\em arXiv preprint arXiv:1606.03798}, 2016.

\bibitem{drost2010model}
B.~Drost, M.~Ulrich, N.~Navab, and S.~Ilic.
\newblock Model globally, match locally: Efficient and robust $3$d object
  recognition.
\newblock In {\em Computer Vision and Pattern Recognition (CVPR), 2010 IEEE
  Conference on}, pages 998--1005. Ieee, 2010.

\bibitem{duda1972use}
R.~O. Duda and P.~E. Hart.
\newblock Use of the hough transformation to detect lines and curves in
  pictures.
\newblock {\em Communications of the ACM}, 15(1):11--15, 1972.

\bibitem{fischler1981random}
M.~A. Fischler and R.~C. Bolles.
\newblock Random sample consensus: a paradigm for model fitting with
  applications to image analysis and automated cartography.
\newblock {\em Communications of the ACM}, 24(6):381--395, 1981.

\bibitem{fragoso2013evsac}
V.~Fragoso, P.~Sen, S.~Rodriguez, and M.~Turk.
\newblock Evsac: accelerating hypotheses generation by modeling matching scores
  with extreme value theory.
\newblock In {\em Computer Vision (ICCV), 2013 IEEE International Conference
  on}, pages 2472--2479. IEEE, 2013.

\bibitem{fragoso2017ansac}
V.~Fragoso, C.~Sweeney, P.~Sen, and M.~Turk.
\newblock Ansac: Adaptive non-minimal sample and consensus.
\newblock In {\em BMVC}, page arXiv:1709.09559, 2017.

\bibitem{holz2015registration}
D.~Holz, A.~E. Ichim, F.~Tombari, R.~B. Rusu, and S.~Behnke.
\newblock Registration with the point cloud library: A modular framework for
  aligning in 3-d.
\newblock {\em IEEE Robotics \& Automation Magazine}, 22(4):110--124, 2015.

\bibitem{hough1959machine}
P.~V. Hough.
\newblock Machine analysis of bubble chamber pictures.
\newblock In {\em International conference on high energy accelerators and
  instrumentation}, volume~73, page~2, 1959.

\bibitem{irschara2009structure}
A.~Irschara, C.~Zach, J.-M. Frahm, and H.~Bischof.
\newblock From structure-from-motion point clouds to fast location recognition.
\newblock In {\em Computer Vision and Pattern Recognition, 2009. CVPR 2009.
  IEEE Conference on}, pages 2599--2606. IEEE, 2009.

\bibitem{kendall2015posenet}
A.~Kendall, M.~Grimes, and R.~Cipolla.
\newblock Posenet: A convolutional network for real-time 6-dof camera
  relocalization.
\newblock In {\em Computer Vision (ICCV), 2015 IEEE International Conference
  on}, pages 2938--2946. IEEE, 2015.

\bibitem{kneip2014opengv}
L.~Kneip and P.~Furgale.
\newblock Opengv: A unified and generalized approach to real-time calibrated
  geometric vision.
\newblock In {\em Robotics and Automation (ICRA), 2014 IEEE International
  Conference on}, pages 1--8. IEEE, 2014.

\bibitem{kneip2011novel}
L.~Kneip, D.~Scaramuzza, and R.~Siegwart.
\newblock A novel parametrization of the perspective-three-point problem for a
  direct computation of absolute camera position and orientation.
\newblock In {\em Computer Vision and Pattern Recognition (CVPR), 2011 IEEE
  Conference on}, pages 2969--2976. IEEE, 2011.

\bibitem{latent_supp}
S.~Korman and R.~Litman.
\newblock Latent ransac supplementary materials.
\newblock \url{arxiv.org/src/1802.07045v2/anc/supp.pdf}, 2018.

\bibitem{lebeda2012fixing}
K.~Lebeda, J.~Matas, and O.~Chum.
\newblock Fixing the locally optimized ransac--full experimental evaluation.
\newblock In {\em British machine vision conference}, pages 1--11. Citeseer,
  2012.

\bibitem{litman2015inverting}
R.~Litman, S.~Korman, A.~Bronstein, and S.~Avidan.
\newblock Inverting ransac: Global model detection via inlier rate estimation.
\newblock In {\em Proceedings of the IEEE Conference on Computer Vision and
  Pattern Recognition}, pages 5243--5251, 2015.

\bibitem{lowe2004distinctive}
D.~G. Lowe.
\newblock Distinctive image features from scale-invariant keypoints.
\newblock {\em International journal of computer vision}, 60(2):91--110, 2004.

\bibitem{mellado2014super}
N.~Mellado, D.~Aiger, and N.~J. Mitra.
\newblock Super 4pcs fast global pointcloud registration via smart indexing.
\newblock In {\em Computer Graphics Forum}, volume~33, pages 205--215. Wiley
  Online Library, 2014.

\bibitem{olsson2008polynomial}
C.~Olsson, O.~Enqvist, and F.~Kahl.
\newblock A polynomial-time bound for matching and registration with outliers.
\newblock In {\em Computer Vision and Pattern Recognition, 2008. CVPR 2008.
  IEEE Conference on}, pages 1--8. IEEE, 2008.

\bibitem{raguram2013usac}
R.~Raguram, O.~Chum, M.~Pollefeys, J.~Matas, and J.-M. Frahm.
\newblock Usac: a universal framework for random sample consensus.
\newblock {\em IEEE transactions on pattern analysis and machine intelligence},
  35(8):2022--2038, 2013.

\bibitem{raguram2011recon}
R.~Raguram and J.-M. Frahm.
\newblock Recon: Scale-adaptive robust estimation via residual consensus.
\newblock In {\em Computer Vision (ICCV), 2011 IEEE International Conference
  on}, pages 1299--1306. IEEE, 2011.

\bibitem{rusu2009fast}
R.~B. Rusu, N.~Blodow, and M.~Beetz.
\newblock Fast point feature histograms (fpfh) for $3$d registration.
\newblock In {\em Robotics and Automation, 2009. ICRA'09. IEEE International
  Conference on}, pages 3212--3217. IEEE, 2009.

\bibitem{shao2003zubud}
H.~Shao, T.~Svoboda, and L.~Van~Gool.
\newblock Zubud - zurich buildings database for image based recognition.
\newblock {\em Computer Vision Lab, Swiss Federal Institute of Technology,
  Switzerland, Tech. Rep}, 260:20, 2003.

\bibitem{tordoff2002guided}
B.~Tordoff and D.~W. Murray.
\newblock Guided sampling and consensus for motion estimation.
\newblock In {\em European conference on computer vision}, pages 82--96.
  Springer, 2002.

\bibitem{vedaldi08vlfeat}
A.~Vedaldi and B.~Fulkerson.
\newblock {VLFeat}: An open and portable librar of computer vision algorithms,
  2008.

\bibitem{wald1973sequential}
A.~Wald.
\newblock {\em Sequential analysis}.
\newblock Courier Corporation, 1973.

\bibitem{xu1990new}
L.~Xu, E.~Oja, and P.~Kultanen.
\newblock A new curve detection method: randomized hough transform (rht).
\newblock {\em Pattern recognition letters}, 11(5):331--338, 1990.

\bibitem{yang2013goicp}
J.~Yang, H.~Li, and Y.~Jia.
\newblock Go-icp: Solving 3d registration efficiently and globally optimally.
\newblock In {\em Proceedings of the IEEE International Conference on Computer
  Vision}, pages 1457--1464, 2013.

\bibitem{zheng2011deterministically}
Y.~Zheng, S.~Sugimoto, and M.~Okutomi.
\newblock Deterministically maximizing feasible subsystem for robust model
  fitting with unit norm constraint.
\newblock In {\em Computer Vision and Pattern Recognition (CVPR), 2011 IEEE
  Conference on}, pages 1825--1832. IEEE, 2011.

\bibitem{zhou2016fast}
Q.-Y. Zhou, J.~Park, and V.~Koltun.
\newblock Fast global registration.
\newblock In {\em European Conference on Computer Vision}, pages 766--782.
  Springer, 2016.

\end{thebibliography}
